\pdfoutput=1

\documentclass[11pt]{article}

\usepackage[]{acl}

\usepackage{times}
\usepackage{latexsym}
\usepackage{booktabs}
\usepackage{multirow}
\usepackage[figuresright]{rotating}
\usepackage{enumitem}
\usepackage{longtable}
\usepackage{courier}
\usepackage{url}
\definecolor{c1}{HTML}{E9F3E3}
\definecolor{c2}{HTML}{F2EAC4}
\definecolor{c3}{HTML}{E6DBF2}

\usepackage[T1]{fontenc}
\usepackage{soul}

\usepackage[utf8]{inputenc}

\usepackage{microtype}
\usepackage{inconsolata}
\usepackage{longtable}
\title{Towards LLM-based Fact Verification on News Claims with a Hierarchical Step-by-Step Prompting Method}

\author{Xuan Zhang \and Wei Gao \\
School of Computing and Information Systems \\ 
Singapore Management University \\
80 Stamford Rd, Singapore 178902 \\
\texttt{xuanzhang.2020@phdcs.smu.edu.sg},~~ \texttt{weigao@smu.edu.sg} }

\begin{document}
\maketitle
\begin{abstract}
While large pre-trained language models (LLMs) have shown their impressive capabilities in various NLP tasks, they are still under-explored in the misinformation domain. In this paper, we examine LLMs with in-context learning (ICL) for news claim verification, and find that only with 4-shot demonstration examples, the performance of several prompting methods can be comparable with previous supervised models. To further boost performance, we introduce a Hierarchical Step-by-Step (HiSS) prompting method which directs LLMs to separate a claim into several subclaims and then verify each of them via multiple questions-answering steps progressively.
Experiment results on two public misinformation datasets show that HiSS prompting outperforms state-of-the-art fully-supervised approach and strong few-shot ICL-enabled baselines.
\end{abstract}

\section{Introduction}
\label{sec:intro}
Misinformation such as fake news often causes confusion or wrong belief because they contain claims that are factually false or inaccurate~\cite{lazer2018science}. 
To combat misinformation in news claims, stakeholders rely on fact-checking practices for claim verification. Fact-checking services online, such as PolitiFact\footnote{\url{https://www.politifact.com/}.} and Snopes\footnote{\url{https://www.snopes.com/}.}) require laborious manual efforts, making it challenging to match the rapid pace of misinformation being produced.

In recent years, deep neural networks-based misinformation detection and fact-checking methods have been studied extensively~\cite{wang2017liar,rashkin2017truth,popat2018declare,ma2019sentence,kotonya2020explainable,atanasova2020generating,yang2022coarse}.
In particular, pre-trained language models (PLMs) like BERT~\cite{kenton2019bert}
have demonstrated superior results and surpassed traditional methods in fake news related benchmarks~\cite{soleimani2020bert,atanasova2020generating,kruengkrai2021multi}, thanks to their strong ability to understand nuanced context for more accurate decision.
Recently, large pre-trained language models (LLMs) with a massive number of parameters, such as GPT-3.5, have shown impressive performances in various downstream tasks~\cite{brown2020language,wei2022chain,zhou2022least,press2022measuring}. But it is basically unclear how well LLMs can perform on fact verification task as this is not at the core of LLM pre-training~\cite{brown2020language,anil2023palm}.

While it is not practical to directly fine-tune most LLMs, \textit{in-context learning} (ICL)~\cite{brown2020language} offers an alternative way to instruct LLMs to learn new tasks via inference only, conditioning on demonstration examples without any gradient updates. 
Properly prompted LLMs can carry out similar steps of logical traces with that in demonstration examples, which is known as Chain-of-Thought (CoT) reasoning~\cite{wei2022chain}. 
This generative reasoning process not only enhances the model's performance on tasks such as arithmetic, commonsense, and symbolic reasoning, but also facilitates the understanding of the underlying rationale behind the results from LLMs.

\begin{figure}[t!]
\centering
\includegraphics[width=0.45\textwidth]{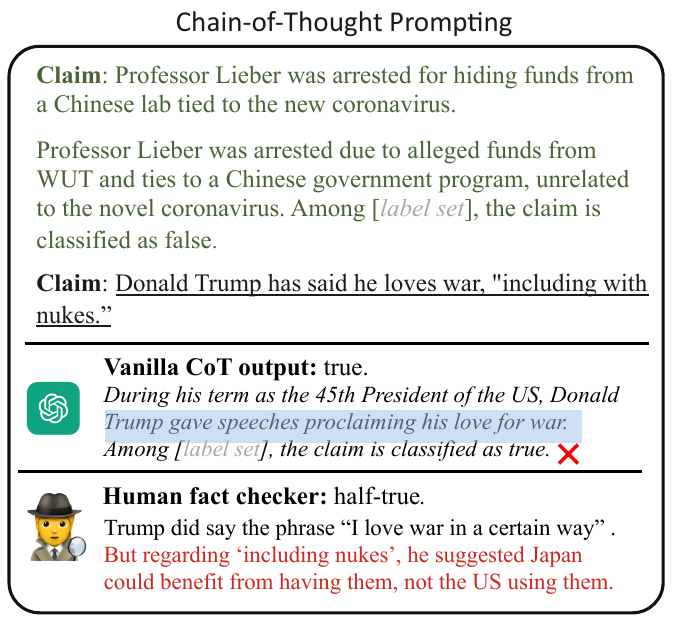} 
\caption{An example of claim verification based on vanilla CoT prompting. The claim (underlined) and CoT (in green) are given as a demonstration. The generated CoT (in italics) leads to an incorrect judgment due to (1) omission of necessary thoughts regarding ``nukes'', and (2) fact hallucination about the war-loving speeches without specific evidence in the generated CoT (in blue).}
\label{demons}
\end{figure}

Previous research has suggested the importance of reasoning in improving the accuracy and explainability of fake news detection~\cite{jin2022towards}. However, leveraging LLM reasoning in the context of fake news related tasks remains under-explored.
In this work, we first evaluate three classical ICL methods, including standard prompting and CoT-based methods for news claim verification. The standard prompting takes in a news claim for LLM to return its factuality judgment on the claim, while CoT additionally generates a series of intermediate verbal reasoning steps in the result. On two fake news benchmark datasets RAWFC~\cite{yang2022coarse} and LIAR~\cite{wang2017liar}, we find that the standard prompting performs comparably well as strong supervised baselines, but the vanilla CoT is worse than the standard prompting, which is counter-intuitive. We found that there are two main issues causing the failure of vanilla CoT, as illustrated in Figure~\ref{demons}: (1) Omission of necessary thoughts -- vanilla CoT tends to ignore some noteworthy parts in the claim, resulting in inaccurate decisions; (2) Fact hallucination\footnote{This type of hallucination is also referred to as the extrinsic hallucination~\cite{bang2023multitask} that cannot be verified with the given source, and the fact-conflicting hallucination~\cite{zhang2023siren} that, more broadly, are not faithful to established world knowledge.} -- When necessary information is not available, the model tends to generate relevant but unreliable ``facts'' on its own, which misleads the final prediction. 
 
To address the issues, we instruct LLMs to decompose a complex claim into smaller subclaims, so that the reasoning follows up with the fine-grained decomposition. This aims to enable a much more thorough examination of the claim, reducing the risk of overlooking necessary details in the claim and enhancing the reasoning effect based on different reasoning chains. This is analogous to breaking down complex questions into subquestions~\cite{press2022measuring} for QA and devising a plan for solving complex tasks into multiple steps~\cite{wang2023plan}.
Additionally, we instruct LLM to employ a search engine for providing up-to-date external information, aiding the model in reasoning and mitigating the hallucination problem.
In light of this, we propose a \textbf{H}ierarchical \textbf{S}tep-by-\textbf{S}tep (\textbf{HiSS}) prompting method, which is composed of two main processes: (1) \textit{Claim Decomposition}, which prompts the LLM to split a complex claim into smaller subclaims.
(2) \textit{Subclaim Verification}, which prompts LLM to verify the subclaim step-by-step employing a search engine to obtain relevant evidence. Our contributions are three-fold:
\begin{itemize}[leftmargin=*]
    \item We investigate the ability of LLMs with ICL for news claim verification. And we find that with only four-shot demonstration examples, LLMs can outperform most of the supervised methods, which indicates LLM is a promising tool to combat misinformation.
    \item We propose a HiSS prompting method to prompt LLM to do fine-grained checking of news claims. Experiments on two public datasets show that HiSS-prompted LLMs outperform traditionally strong fully-supervised models with an improvement of 4.95\% on average in macro-average F1 and set a new state-of-the-art for few-shot news claim verification\footnote{Code and prompts data is available at \url{https://github.com/jadeCurl/HiSS}.}.
    \item Compared with previous methods, our HiSS-prompted LLMs provide superior explanations, which are more fine-grained and easier to follow based on automatic and human evaluation.
\end{itemize}

\section{Related Work}
\subsection{Explainable Fake News Detection}
Existing research on explainable fake news detection is mainly focused on generating explanations from input evidence. These approaches include generating human-comprehensible explanations for candidate facts based on background knowledge
encoded in the form of Horn clauses~\cite{gad2019exfakt}, as well as using attention-based models to highlight relevant factual words~\cite{popat2018declare}, news attributes~\cite{yang2019xfake} and suspicious users~\cite{lu2020gcan}. Such an approach is based on general deep neural networks and knowledge base instead of language models. 

Later, 
\citet{atanasova2020generating} and \citet{kotonya2020explainable} propose directly producing veracity explanations based on extractive and abstractive summarization. However, these methods predominantly generate explanations by summarizing fact-checking articles. While such an approach can somewhat explain fact-checking decisions following human thoughts written in the articles, it does not reason based on raw evidence to form the thoughts for drawing conclusions, which should be the core of fact verification.

\subsection{Fact Verification with Language Models}
Previous research has utilized PLMs (e.g., BERT and BART) in fake news related tasks. For example, \citet{lee2020language} directly uses the internal knowledge implicitly stored as PLMs' parameters for fact verification. \citet{lewis2020retrieval} proposes a retrieval-augmented approach to endow language models with document retrieval capability, which was applied for selecting relevant evidence in fact extraction and verification. Instead of using language models to provide evidence only, \citet{lee2021towards} utilizes LLMs such as GPT-2~\cite{radford2019language} and their few-shot capability to assess the claim's factuality based on the perplexity of evidence-conditioned claim generation. 

Research on utilizing the reasoning capabilities of LLMs, such as CoT-based reasoning, in the misinformation domain is still limited. 
Recent works~\cite{press2022measuring,yao2023react,jiang2023active} find that combining LLM's reasoning capability with accessibility to external knowledge is helpful to many reasoning-intensive NLP tasks including HotpotQA~\cite{yang2018hotpotqa} and FEVER~\cite{thorne2018fever}. In contrast to existing works, our research is motivated by the counter-intuitive observation that CoT under-performs the standard prompting in news claim verification, and explores how to better elicit LLMs to mitigate two salient issues of LLMs in this task. We focus on the verification of real-world news claims, which could be more temporally dynamic and sensitive than FEVER type of claims, necessitating the model to access up-to-date knowledge.

\section{Our HiSS Prompting Method}

\begin{figure*}[thb!]
\centering
\centering 
\includegraphics[width=1\textwidth]{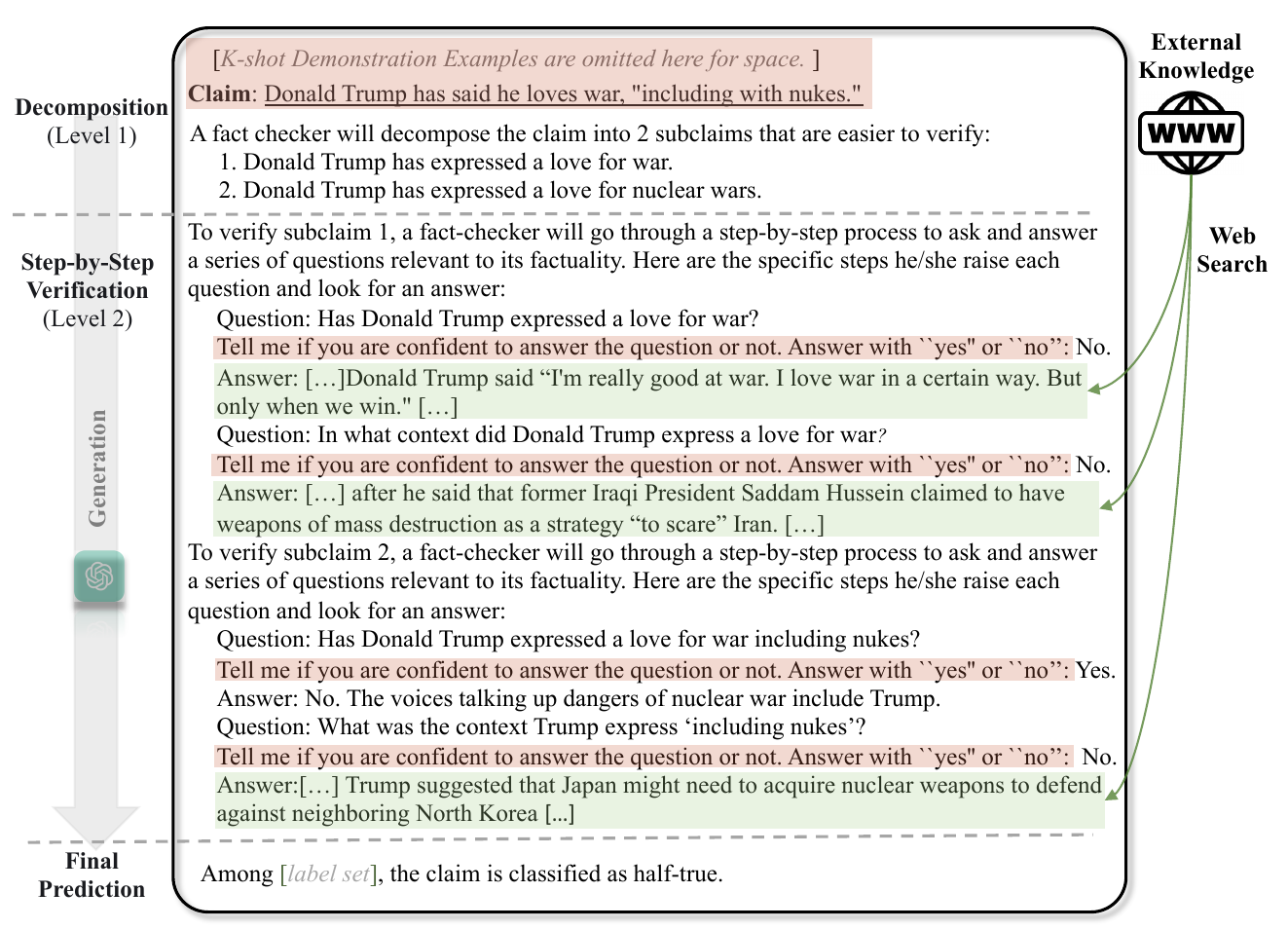} 
\caption{Overview of the proposed HiSS model: Original human inputs are in red background, LLM directly generated text is in white, and answers generated based on search results are in green. We start by providing a few-shot demonstration, followed by appending the claim to be checked (underlined). HiSS prompts the LLM to (1) decompose the claim into subclaims; (2) verify each subclaim step-by-step via raising and answering a series of questions. For each question, we prompt LLM to assess if it is confident to answer it or not, and if not, we input the question to a web search engine. The search results are then inserted back into the ongoing prompt to continue the verification process; (3) generate the final prediction. The detailed demonstrations are omitted in this illustration for space which can be found in Table~\ref{tbl:hiss} and Table~\ref{tbl:hiss_2} in Appendix~\ref{app:prompt} .}
\label{framework}
\end{figure*}

In this section, we address the two main issues of LLMs observed in the news claim verification task, i.e., 1) Omission of necessary thoughts and 2) Fact hallucination. We will first raise our specific research questions, and then present our HiSS prompting method.

\subsection{Research Questions}
For the omission of necessary thoughts, the basic research question we need to address would be: 
\begin{itemize}
\item \emph{How to instruct LLMs not to overlook any crucial points of a claim in its CoT?}
\end{itemize}
The context of real-world claims could be complex and deep. For example, the seemingly easy claim \emph{Donald Trump has said he loves war, ``including with nukes''} is actually quite intricate, as it not only explicitly states Trump's declaration of love for both regular and nuclear wars, but also implies that in order to verify the statement is factual or not, one has to examine whether and in what circumstances he has expressed such passion on both types of wars. Therefore, we propose to prompt LLMs to thoroughly generate all explicit and implicit points that are check-worthy given a claim. 

Hallucination is an intrinsic and fundamental problem of LLMs~\cite{ji2023survey,bang2023multitask}. We address it by providing relevant and up-to-date contextual information to LLM as external knowledge, assuming that hallucinations most likely result from the lack of knowledge on the necessary context~\cite{bang2023multitask}. Our specific research question would be:

\begin{itemize}
\item \emph{How can we determine when external knowledge is needed during the verification and assist LLM in acquiring the necessary knowledge to mitigate fact hallucination?}
\end{itemize}
While the decomposition can prompt LLM to raise fine-grained questions, the model may make up responses when background information is lacking. For instance, if the model is unaware of the specific contexts of Trump's wording on ``war''\footnote{The comment regarding Trump's ``love'' of war comes from his speech in Iowa on Nov. 12, 2015. In the speech, Trump theorized that former Iraqi leader Saddam Hussein feigned having weapons of mass destruction to scare Iran, before briefly sidetracking into his feelings on war generally: ``This is the Trump theory on war,'' he said. ``But I'm good at war. I've had a lot of wars of my own. I'm really good at war. I love war in a certain way. But only when we win.''} and ``nukes''\footnote{Trump made his comments about ``nukes'' in an April 3 interview with Fox News Sunday’s Chris Wallace. Wallace was asking Trump about his suggestion that Japan might be better off with nuclear weapons. Trump suggested that Japan might need to acquire nuclear
weapons to defend against neighboring North Korea. It's worth noting that the comment wasn’t about the United States using nuclear weapons, but about his belief that Japan might be better off if it had nuclear weapons.}, it can lead to factually inaccurate answer, such as ``During his term as the 45th President of the US, Donald Trump gave speeches proclaiming his love for war''.

In the following subsections, we will describe our Hierarchical Step-by-Step (HiSS) prompting method. As shown in Figure~\ref{framework}, HiSS involves three processes: (1) \textit{Claim decomposition}, (2) \textit{Subclaim step-by-step verification}, and (3) \textit{Final prediction}.

\subsection{Claim Decomposition}

At the first level of HiSS, we focus on instructing the model to capture all the explicit points in the original claim and decompose them into subclaims.  This level aligns with previous studies~\cite{ousidhoum2022varifocal,fan2020generating} in fact-checking, which found that segmenting the original claim by identifying entities or focal points can facilitate human fact-checkers in making informed judgments. However, these models require the manual collection of datasets for training, whereas our method prompts LLM to do the decomposition guided by only a few demonstration examples.

Speficially, LLM is prompted with $K$-shot ($K$ is a hyperparameter) demonstration examples (see Table~\ref{tbl:hiss} and Table~\ref{tbl:hiss_2} in Appendix~\ref{app:prompt} for details) that serve to illustrate the entire verification process, followed by the test claim to be checked, as shown in the Level 1 in Figure~\ref{framework}. 
The demonstration examples exhibit to LLM how to break down a claim $c_i$ into a series of subclaims $[s_{i1}, s_{i2}, \cdots, s_{iN_i}]$ that cover all check-worthy points explicitly expressed. 
The demonstration examples vary in their complexity, with some simple claims not undergoing deep decomposition and more complex claims being decomposed into a few more subclaims. The LLM presumably follows the demonstrated decomposition approach in accordance with the complexity of the input claim $c_i$. Therefore, $N_i$ is determined by LLM automatically. Figure~\ref{framework} illustrates that LLM decomposes the test claim into two subclaims.

\subsection{Subclaim-level Step-by-Step Verification}
\label{sec:sbs}
In the second level, LLM  individually verifies each subclaim obtained from Level 1. Underlying the explicit points conveyed in each subclaim can be a few implicit points that are not expressed but need further scrutinization in one way or the other. For example, ``Did Trump really say he loves war?'', ``What is his exact wording?'', ``In what context did he express it?'', etc. for the first subsclaim ``Donald Trump has expressed a love for war''.

Specifically, we leverage the reasoning capability of LLM to delve deeper into the underlying information needed to validate each subclaim $s_{ij}$ by generating a series of probing questions $\{q_{ij}^m\}$, each $q_{ij}^m$ corresponding to an implicit point. Similarly, the number of probing questions of each subclaim is determined by LLM automatically with reference to the demonstration example.
We adopt a progressive approach to generate the questions. This allows us to adjust the subsequent question generation based on the answers to previous questions and the acquired context information on the chain. As a result, the generated questions become more targeted and in-depth, facilitating a comprehensive analysis of the subclaims.

Once a question $q_{ij}^m$ is generated, the next step is to elicit the corresponding answer $a_{ij}^m$ from LLM. Recent works have found that providing LLMs with access to external knowledge can lead to notable improvements~\cite{yao2023react,jiang2023active}. An important consideration is how to prompt LLM to automatically decide when it needs to consult with an external knowledge source (e.g., web search engine), to mitigate fact hallucination. It is hypothesized that LLM can be prompted to assess its own confidence in answering a question, so that we can acquire relevant external information to aid it when it lacks confidence. We resort to Google Search as an external source. 

Specifically, LLM follows the specific format of demonstration examples to generate questions: it starts with the prefix ``\texttt{Question:}'' and presents the generated question $q_{ij}^m$, followed by ``\texttt{Tell me if you are confident ...}''. We control the model to pause at the end of $q_{ij}^m$ by setting the phrase ``\texttt{Tell me if you are confident}'' as the stop sequence\footnote{The ``stop sequence'' mechanism is a setting provided by the OpenAI API (\url{https://help.openai.com/en/articles/5072263-how-do-i-use-stop-sequences}). When a specific word or phrase is set as a ``stop sequence'', the model will halt its generation upon encountering that word or phrase, allowing users to control the length or content of the generated output. }. This aims to facilitate 1) extracting the text of $q_{ij}^m$, and 2) probing the LLM to assess its confidence in answering the question without additional information. During its pause, we append the following instruction: 
\texttt{Tell me if you are confident to answer the question or not. Answer with ``yes'' or ``no'':}, and set the stop sequence to `no'. This means that if the LLM responds with `no', the model will cease to further generate an answer for $q^m_{ij}$, but wait for us to input $q_{ij}^m$ into Google Search API\footnote{\url{https://serpapi.com}.} to obtain top search results\footnote{Search results from fact-checking websites are filtered to avoid ground-truth leakage. Specifically, we remove the search results with URLs containing keywords such as ``fact check'', and ``fact-checking'' since the URL of fact-checking websites and fact-check articles on mainstream media, e.g., NY Times (\url{https://www.nytimes.com/spotlight/fact-checks}.), typically contain such keywords. After filtering, we choose the top-one snippet from the search result to feed into the LLM.}, so that we can feed them into the LLM for it to generate the answer $a_{ij}^m$. 
However, if the LLM responds with ``\texttt{yes}'', the LLM does not halt and proceeds to generate the answer $a_{ij}^m$ to the question. Following the specific format of the demonstration example, after a prior question is addressed, the LLM continues to generate the subsequent question until it ceases to produce any more questions, transitioning then to the final prediction phase.

\subsection{Final Prediction} Once all the subclaims have been verified, the LLM can make a final prediction. At this point, it outputs ``\texttt{Among [\textit{label set}], the claim is classified as}'' before providing the final answer, where \texttt{[\textit{label set}]} is substituted with the actual label set for a specific dataset. This facilitates the parsing of the final prediction, as the predicted class label will appear after the word ``\texttt{as}'' in the last output line.

\section{Experiments and Results}
\subsection{Experimental Setup}
\begin{table}[t!]
\centering

\begin{tabular}{lrrrrrr}
\toprule

& \multicolumn{2}{c}{\textbf{RAWFC}}&\multicolumn{2}{c}{\textbf{LIAR}}\\

 \cmidrule(lr){2-3} \cmidrule(lr){4-5}
 & \textbf{Val.} & \textbf{Test}& \textbf{Val.} & \textbf{Test}\\
 \midrule
Claim & 200 & 200& 1274 & 1,251  \\
~~~\# true & 67 & 67& 169 & 205  \\
~~~\# mostly-true&-&- & 251 & 238  \\
~~~\# half-true & 66 & 66 & 244 & 263  \\
~~~\# barely-true&-&- & 236 & 210  \\
~~~\# false& 67 & 67 & 259 & 249 \\
~~~\# pants-fire&-&- & 115 & 86 \\
\bottomrule

\end{tabular}
\caption{Datasets statistics.}
\label{tbl_datasta}

\end{table}
We conducted experiments on two standard English fake news datasets: 1)  \textbf{RAWFC}~\cite{yang2022coarse} contains gold labels based on Snopes fact-check articles and follows a three-class classification scheme (True/False/Half); 2) \textbf{LIAR}~\cite{wang2017liar} contains gold labels based on PolitiFact articles with six classes (True/Mostly-true/Half-true/Barely-true/False/Pants-fire). Different from FEVER~\cite{thorne2018fever} which uses manually synthesized claims from  Wikipedia articles, the claims in these two datasets are based on real-world news. 
Table~\ref{tbl_datasta} displays the statistics of datasets. We use the provided valid-test split of both datasets. The few-shot demonstration examples are randomly selected from the training set.

Following~\citet{yang2022coarse}, we use macro-average precision ($P$), recall ($R$), and $F_1$ ($F_1=\frac{2 R P}{R+P}$) scores as the metrics for evaluation.

\begin{table*}[t]
\centering

\begin{tabular}{lllllll}
\toprule

\multirow{2}*{\textbf{Model}}& \multicolumn{3}{c}{\textbf{RAWFC}}&\multicolumn{3}{c}{\textbf{LIAR}}\\ 
\cmidrule(lr){2-4} \cmidrule(lr){5-7} 
&$P(\%)$ & $R(\%)$ & $F_1(\%)$&$P(\%)$ & $R(\%)$ & $F_1(\%)$\\\midrule
\multicolumn{7}{l}{\textit{\textbf{Fully Supervised Models}}}\\

CNN~\cite{wang2017liar}&38.8 & 38.5 & 38.6&22.6 & 22.4 & 22.5\\
RNN~\cite{rashkin2017truth}&41.4 & 42.1 & 41.7&24.4 & 21.2 & 22.7\\
DeClarE$^{\dagger}$~\cite{popat2018declare}&43.4 & 43.5 & 43.4&22.9 & 20.6 & 21.7\\
SentHAN$^{\dagger}$~\cite{ma2019sentence}&45.7 & 45.5 & 45.6&22.6 & 20.0 & 21.2\\
SBERT$^{\diamondsuit}$~\cite{kotonya2020explainable}&51.1 & 46.0 & 48.4&24.1 & 22.1 & 23.1\\
GenFE$^{\diamondsuit}$~\cite{atanasova2020generating}&44.3 & 44.8 & 44.5&28.0 & 26.2 & 27.1\\
CofCED$^{\dagger}$~\cite{yang2022coarse}&53.0 & 51.0 & 52.0&29.5 & 29.6 & 29.5\\
\midrule

\multicolumn{7}{l}{\textit{\textbf{Few-shot Models \textit{w/} }}\textbf{GPT3.5}}\\

Standard Prompt~\cite{brown2020language}&48.5  &   48.5   &   48.5&29.1   &   25.1    &  27.0\\
Vanilla CoT~\cite{wei2022chain}&42.4    & 46.6 & 44.4&22.6 & 24.2 & 23.7  \\
Search-Augmented CoT$^{\dagger}$&47.2&51.4&49.2&27.5 &23.6&25.4\\
ReAct$^{\dagger}$~\cite{yao2023react}&51.2&48.5&49.8&33.2&29.0&31.0   \\
HiSS$^{\dagger}$~(ours)&\textbf{53.4}&    \textbf{54.4}$^*$ &  \textbf{53.9}$^*$   &\textbf{46.8}$^*$ &     \textbf{31.3}$^*$ &    \textbf{37.5}$^*$\\
\bottomrule
\end{tabular}

\caption{Experimental results of claim verification. Bold denotes the best performance. $^*$ means significantly better than the previous SoTA (CofCED) with $p<0.01$. $^{\dagger}$ uses external information obtained via search engines. $^{\diamondsuit}$ uses gold evidence from fact-check reports. Results of fully supervised models are quoted from~\cite{yang2022coarse}.}
\label{tbl:main}
\end{table*}

\paragraph{Supervised baselines.} We compare with seven strong supervised models in claim verification: 1) CNN~\cite{wang2017liar} uses a convolutional neural model to integrate claim information and available metadata features (e.g. subject, speaker, and party) to get the prediction; 
2) RNN~\cite{rashkin2017truth} uses recurrent neural networks to learn representation from word sequences of the claim.
3) DeClarE~\cite{popat2018declare} considers word embedding from both the claim and searched external information as evidence. 
4) SentHAN~\cite{ma2019sentence} proposes a hierarchical attention network to represent external evidence as well as their semantic relatedness with the claim. 
5) SBERT-FC~\cite{kotonya2020explainable} uses Sentence-BERT to encode both claim and evidence for classification. 
6) GenFE~\cite{atanasova2020generating} predicts fact-check results and generates explanations in the multi-task setup. 
7) CofCED~\cite{yang2022coarse} uses a hierarchical encoder for text representation and two coarse-to-fine cascaded selectors to extract key evidence for news claim verification.

\paragraph{Few-shot baselines.} We employ the following few-shot baselines for comparison: 

1) Standard Prompting~\cite{brown2020language} directly asks the LLM to determine the class label of the claim. 
2) Vanilla CoT Prompting~\cite{wei2022chain} asks the LLM to output a thought chain before outputting the class label of the claim. The demonstration examples for both standard prompting and vanilla CoT are shown in Table~\ref{tbl:sp} and~\ref{tbl:cot}, respectively (see Appendix~\ref{app:prompt}). 3) Search-Augmented CoT Prompting: To compare with the baselines that can access extra knowledge, we augment vanilla CoT with search engine by using the claim as query to retrieve the background information, and then let the LLM output thought chain and class label based on the information retrieved.
4) ReAct Prompting~\cite{yao2023react} is a variant of CoT that 
explores the use of LLMs to generate both reasoning traces and task-specific actions (e.g., search) in an interleaved manner.
For a fair comparison, we employ the same demonstration examples and search engine across the different systems.

\paragraph{Implementation Details}
  To ensure reproducibility, we generate outputs using greedy decoding by setting the temperature to $0$. We also freeze the search results for the same queries involved in the experiments for fair comparison, as search engine results can potentially change over time across different runs.
We utilize the GPT-3.5 series API \texttt{text-davinci-003}\footnote{\url{https://platform.openai.com/docs/models/gpt-3-5}.} as a backbone LLM.
Following~\citet{wei2022chain}, we tune the hyperparameter of the shot number within $\{1, 2, 4, 6, 8\}$ on the validation set, and find that the model achieved the best results with $4$ demonstration examples. Therefore, we set $K=4$ throughout the experimentation.

\subsection{Results of Claim Veracity Classification}

Table~\ref{tbl:main} summarizes the performance of verification, and we have the following findings:

\begin{itemize}[leftmargin=*]

\item \textbf{HiSS with LLM is comparable with or even better than the fully supervised SoTA.} As shown in Table~\ref{tbl:main}, HiSS outperforms previous SoTA (i.e., CofCED) by 1.9\% and 8\% in F1 on RAWFC and LIAR, respectively.
This indicates that few-shot ICL is promising for news claim verification, owing to the capabilities of LLM that benefit from its parameterized knowledge with a tremendous amount of facts.
In addition, HiSS is specially designed to better elicit LLM and guide it through a step-by-step examination of the claim, covering both explicit and implicit aspects and allowing for more comprehensive and thorough verification. Moreover, HiSS enables evidence acquisition via web search when necessary, mitigating the risk of hallucination.
\item \textbf{The performance of few-shot ICL methods varies.} Despite utilizing the same backbone, HiSS surpasses standard prompting, vanilla CoT, and ReAct by 7.95\%, 11.65\%, and 5.3\% in F1 on average, respectively. This observation highlights the importance of specific methods prompting LLM for news claim verification. After conducting an in-depth error analysis on 40 randomly selected samples for vanilla CoT, ReAct, and HiSS\footnote{We omit standard prompting as it directly outputs the final prediction without providing intermediate or reasoning steps.}, as shown in Table~\ref{tbl:error}, we classify the errors observed in the verification traces into two categories: (1) \textit{fact hallucination} and (2) \textit{omission of necessary thoughts}. We find that vanilla CoT exhibits substantial issues of both hallucination and thought omission. 
Although the Search-Augmented CoT improves its performance, it still falls short of meeting the standard prompting method. This suggests that using the original claim as a search query may end up with insufficiently detailed and informative search results, which explains its subpar performance.
In contrast, ReAct, with its ability to autonomously generate search queries and access external knowledge, effectively mitigates failures caused by hallucinations. However, it encounters challenges of thought omission as it may ignore noteworthy points of a claim due to the lack of claim decomposition and a fine-grained step-by-step process.
Our HiSS prompting method instead effectively addresses both issues, thanks to its ability to cover both explicit and implicit points of the claim to get checked and the ability to seek necessary external knowledge supported by the search engine.
\end{itemize}

\begin{table}[t]
\centering
\begin{tabular}{lcccc}
\toprule
Error Types&CoT&ReAct&HiSS\\
\midrule
Fact Hallucination&43\%&28\%&5\%\\
Thoughts Omission&60\%&53\%&13\%\\
\bottomrule
\end{tabular}

\caption{Distribution of errors based on 40 examples from RAWFC, where Vanilla CoT, ReAct, and HiSS give incorrect verification results.}
\label{tbl:error}
\end{table}

\subsection{Ablation Study}
\label{sec:ana}

\begin{figure}[t!]
  \centering
  \includegraphics[width=0.5\textwidth]{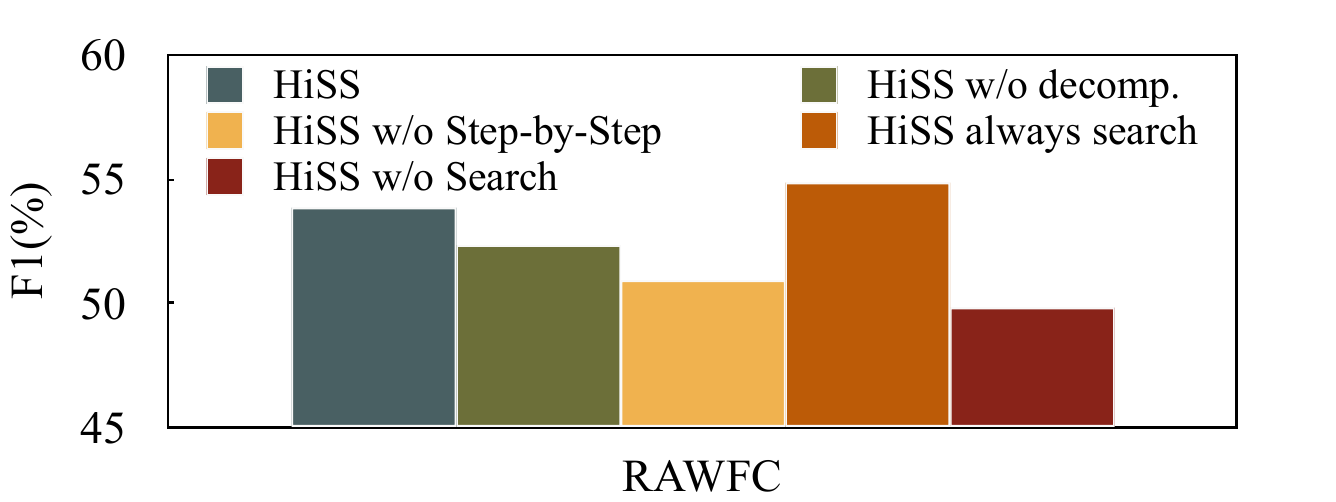}
  \caption{Ablation results on RAWFC dataset.}
  \label{fig:ablation}
  
\end{figure}

To analyze the impact of different configurations of HiSS, we conducted an ablation analysis on RAWFC as shown in Figure~\ref{fig:ablation}. 

\paragraph{Effect of claim decomposition:}
Firstly, we consider HiSS without claim decomposition, where we directly pose probing questions based on the original claim, bypassing the claim decomposition step while keeping the step-by-step process. In this setting, the performance of HiSS decreases by 1.5\%. This result demonstrates that claim decomposition, which separates the claim based on explicit points, is helpful in improving the final predictions.

\paragraph{Effect of subclaim step-by-step verification:}
Next, we conduct an ablation study on the step-by-step verification process for each subclaim. Instead of generating probing questions, we let LLM directly verify subclaims by searching for relevant background information. Subsequently, the LLM made predictions based on the subclaims and the retrieved information. Notably, this modification resulted in a 2.9\% performance drop, underscoring the importance of employing the subclaim step-by-step verification approach to address the implicit points associated with each subclaim.

\paragraph{Effect of strategy using search:}
We compared three approaches to explore the effect of different strategies on using the search function or not and how it is used: 1) \textit{HiSS w/o search}, which relies solely on the internal knowledge from LLM, 2) \textit{HiSS always search}, which always queries the search engine to access the external knowledge, and 3) \textit{HiSS}, which lets LLM self-decide whether to use the search results in each step based on its own confidence (see Section~\ref{sec:sbs}).

As expected, the performance of \textit{HiSS w/o search} is poor which achieves only an F1 of 49.8\%, indicating that reliance solely on LLM's internal knowledge is unreliable and insufficient. An interesting finding is that \textit{HiSS} prompted to decide whether to leverage search results or not based on the self-confidence of LLM achieves an F1 of 54.4\%, which is just slightly worse than the \textit{HiSS always search} by 1.0\%. Our further inspection reveals that out of the 200 test claims on RAWFC, a total of 934 questions are generated, and LLM flags 690 of them as being confident to answer.
This indicates that in cases where the model is confident, external knowledge from the search engine can only marginally improve its performance, as the model is capable of providing accurate answers. In contrast, for the cases where the model lacks confidence, leveraging search results can enhance its performance much more greatly.  Assuming that we can basically trust the factuality of search results from the web, this suggests that the model demonstrates a reasonably good estimation of its own confidence.

\subsection{Human Evaluation}
We conduct a human evaluation to study the explanation quality of three different types of explanations: Gold justification given by human journalists, explanations generated by the strongest supervised explainable model CofCED, and the reasoning trajectory generated from the HiSS method. We ask three English-speaking judges to rate these explanations with scores of 1, 2, and 3 (higher is better) according to the following criteria:
\begin{itemize}[leftmargin=*]
    \item\textbf{Coverage}. The explanation and reasoning does not miss any important points that contribute to the check.
    \item\textbf{Non-redundancy}. The explanation and reasoning provided only includes relevant information that is necessary for understanding the claim and fact-checking it, without any redundant or repeated information.
    \item\textbf{Readability}. The explanation and reasoning is straightforward and simple to read.
    \item\textbf{Overall}. The overall quality of the generated explanation and reasoning.
\end{itemize}
We randomly sample 34 claims from the LIAR test set. Three annotators rate them independently. We compute Krippendorff’s $\alpha$ inter-annotator agreement (IAA)~\cite{hayes2007answering} and get $0.36$ for coverage, $0.42$ for non-redundancy, $0.30$ for readability and $0.38$ for overall.

\begin{table}[t!]
\centering
\begin{tabular}{lccc}
\toprule
& \multicolumn{3}{c}{\textbf{RAWFC}}\\ 
\cmidrule(lr){2-4} 
&Gold&CofCED&HiSS\\
 \midrule

Readability&\textbf{2.75}&1.63&\ul{2.44}\\
Coverage&\textbf{2.65}&1.99&\ul{2.63}\\
Non-redundancy&\textbf{2.72}&1.28&\ul{2.25}\\
Overall&\textbf{2.69}&1.74&\ul{2.54}\\

\bottomrule

\end{tabular}
\caption{Average human ratings on explanations of verification for the claims in the RAWFC dataset. Gold, CofCED, and HiSS correspond to the explanations produced by human journalists, CofCED and HiSS, respectively. A higher score means a better explanation. The highest score is in bold, and the second is underlined.}
\label{tbl_human}
\end{table}

Table~\ref{tbl_human} shows the averaged scores of human evaluation. We find that the gold explanations are slightly better than HiSS-based explanations, while the state-of-the-art automatic explainable claim verification model CofCED is the worst. In particular, for the coverage criteria, HiSS can elicit explanations that are on par with the human-written ones. This explains that our HiSS elicits GPT-3.5 to generate more fine-grained checking points and steps. In addition, the non-redundancy score is relatively lower, since GPT-3.5 may generate repeated subclaims. We conjecture that this may be due to the intrinsic problem of greedy sampling of language models~\cite{holtzman2019curious}.

\section{Conclusion and Future Work}
In this paper, we study different prompting methods for using LLMs in news claim verification. We introduce a hierarchical step-by-step (HiSS) method that prompts LLM to perform the verification in fine-grained steps, aiming to mitigate the omission of thoughts and fact hallucination.
Validated on two public datasets, HiSS prompting improves the performance of LLMs on the task over fully-supervised SoTA models and its strong few-shot ICL-based counterparts. HiSS prompted explanations show superior explainability in their coverage and readability.

In the future, we will build a conversational fact-checking model based on LLMs which can be user-friendly and incorporate human fact-checkers in the loop.

\section{Limitations}
Despite the promising performance of LLMs based on few-shot ICL, fact verification is a challenging research problem given the fact that performance scores are still quite low in general. There are a few limitations.
Firstly, in this work, we highlight that all the baselines and our proposed method solely rely on textual information. We focus on an unimodal approach utilizing language models and do not consider the potential assistance from other modalities, such as images and videos, for this task. Although the exploration of multimodal approaches has gradually drawn some research attention~\cite{wang2018eann,silva2021embracing,bu2023online}, it falls outside the scope of our current work. 

Meanwhile,  the scope of this study is limited to the verification of news claims, which represents only a subset of the broader issue of misinformation. Misinformation encompasses a wide range of false or misleading information, including rumors, fake news articles, and spams~\cite{wu2019misinformation}. While our focus was specifically on news claims, future research could explore the detection and mitigation of misinformation in other formats.

Further, our proposed prompting method heavily relies on the capabilities of backbone LLMs, which can come with substantial computational costs. Our method leverages the advancements in multi-step reasoning exhibited by these LLMs, necessitating high-performance expectations. However, it is worth noting that most state-of-the-art LLMs are currently not open-source and only available as services. For instance, GPT-3.5 can only be accessed via API. The reliance on such LLMs makes deep model control infeasible, and the need for API access poses challenges in terms of cost.

Finally, while our approach leverages search engines to mitigate the fact hallucination issue in LLMs, it operates under the assumption that pertinent information is readily accessible through web search. However, not all information is indexed or available in search engines. For instance, if someone claims to have witnessed a rare meteorological phenomenon in a small town, such event might not be reported on major news websites or databases. Such firsthand, non-digitized accounts might be retrieved or fact-checked. This underscores the limitation in relying solely on search engines as a primary source of external knowledge for fact-checking with LLMs. Another limitation of our method lies in the claims that are beyond established world knowledge when necessary relevant knowledge is not complete or even not available. This necessitates the model's ability to infer novel knowledge by formulating and subsequently validating appropriate hypotheses, a task that remains beyond the capabilities of existing technologies.

\section*{Acknowledgement}
We thank the anonymous reviewers for their helpful comments during the review of this paper.

\newpage

\bibliography{anthology,custom}

\begin{thebibliography}{40}
\expandafter\ifx\csname natexlab\endcsname\relax\def\natexlab#1{#1}\fi

\bibitem[{Anil et~al.(2023)Anil, Dai, Firat, Johnson, Lepikhin, Passos,
  Shakeri, Taropa, Bailey, Chen, Chu, Clark, Shafey, Huang, Meier-Hellstern,
  Mishra, Moreira, Omernick, Robinson, Ruder, Tay, Xiao, Xu, Zhang, Abrego,
  Ahn, Austin, Barham, Botha, Bradbury, Brahma, Brooks, Catasta, Cheng, Cherry,
  Choquette-Choo, Chowdhery, Crepy, Dave, Dehghani, Dev, Devlin, Díaz, Du,
  Dyer, Feinberg, Feng, Fienber, Freitag, Garcia, Gehrmann, Gonzalez, Gur-Ari,
  Hand, Hashemi, Hou, Howland, Hu, Hui, Hurwitz, Isard, Ittycheriah, Jagielski,
  Jia, Kenealy, Krikun, Kudugunta, Lan, Lee, Lee, Li, Li, Li, Li, Li, Lim, Lin,
  Liu, Liu, Maggioni, Mahendru, Maynez, Misra, Moussalem, Nado, Nham, Ni,
  Nystrom, Parrish, Pellat, Polacek, Polozov, Pope, Qiao, Reif, Richter, Riley,
  Ros, Roy, Saeta, Samuel, Shelby, Slone, Smilkov, So, Sohn, Tokumine, Valter,
  Vasudevan, Vodrahalli, Wang, Wang, Wang, Wang, Wieting, Wu, Xu, Xu, Xue, Yin,
  Yu, Zhang, Zheng, Zheng, Zhou, Zhou, Petrov, and Wu}]{anil2023palm}
Rohan Anil, Andrew~M. Dai, Orhan Firat, Melvin Johnson, Dmitry Lepikhin,
  Alexandre Passos, Siamak Shakeri, Emanuel Taropa, Paige Bailey, Zhifeng Chen,
  Eric Chu, Jonathan~H. Clark, Laurent~El Shafey, Yanping Huang, Kathy
  Meier-Hellstern, Gaurav Mishra, Erica Moreira, Mark Omernick, Kevin Robinson,
  Sebastian Ruder, Yi~Tay, Kefan Xiao, Yuanzhong Xu, Yujing Zhang,
  Gustavo~Hernandez Abrego, Junwhan Ahn, Jacob Austin, Paul Barham, Jan Botha,
  James Bradbury, Siddhartha Brahma, Kevin Brooks, Michele Catasta, Yong Cheng,
  Colin Cherry, Christopher~A. Choquette-Choo, Aakanksha Chowdhery, Clément
  Crepy, Shachi Dave, Mostafa Dehghani, Sunipa Dev, Jacob Devlin, Mark Díaz,
  Nan Du, Ethan Dyer, Vlad Feinberg, Fangxiaoyu Feng, Vlad Fienber, Markus
  Freitag, Xavier Garcia, Sebastian Gehrmann, Lucas Gonzalez, Guy Gur-Ari,
  Steven Hand, Hadi Hashemi, Le~Hou, Joshua Howland, Andrea Hu, Jeffrey Hui,
  Jeremy Hurwitz, Michael Isard, Abe Ittycheriah, Matthew Jagielski, Wenhao
  Jia, Kathleen Kenealy, Maxim Krikun, Sneha Kudugunta, Chang Lan, Katherine
  Lee, Benjamin Lee, Eric Li, Music Li, Wei Li, YaGuang Li, Jian Li, Hyeontaek
  Lim, Hanzhao Lin, Zhongtao Liu, Frederick Liu, Marcello Maggioni, Aroma
  Mahendru, Joshua Maynez, Vedant Misra, Maysam Moussalem, Zachary Nado, John
  Nham, Eric Ni, Andrew Nystrom, Alicia Parrish, Marie Pellat, Martin Polacek,
  Alex Polozov, Reiner Pope, Siyuan Qiao, Emily Reif, Bryan Richter, Parker
  Riley, Alex~Castro Ros, Aurko Roy, Brennan Saeta, Rajkumar Samuel, Renee
  Shelby, Ambrose Slone, Daniel Smilkov, David~R. So, Daniel Sohn, Simon
  Tokumine, Dasha Valter, Vijay Vasudevan, Kiran Vodrahalli, Xuezhi Wang,
  Pidong Wang, Zirui Wang, Tao Wang, John Wieting, Yuhuai Wu, Kelvin Xu, Yunhan
  Xu, Linting Xue, Pengcheng Yin, Jiahui Yu, Qiao Zhang, Steven Zheng,
  Ce~Zheng, Weikang Zhou, Denny Zhou, Slav Petrov, and Yonghui Wu. 2023.
\newblock \href {http://arxiv.org/abs/2305.10403} {Palm 2 technical report}.

\bibitem[{Atanasova et~al.(2020)Atanasova, Simonsen, Lioma, and
  Augenstein}]{atanasova2020generating}
Pepa Atanasova, Jakob~Grue Simonsen, Christina Lioma, and Isabelle Augenstein.
  2020.
\newblock Generating fact checking explanations.
\newblock In \emph{ACL}, pages 7352--7364.

\bibitem[{Bang et~al.(2023)Bang, Cahyawijaya, Lee, Dai, Su, Wilie, Lovenia, Ji,
  Yu, Chung et~al.}]{bang2023multitask}
Yejin Bang, Samuel Cahyawijaya, Nayeon Lee, Wenliang Dai, Dan Su, Bryan Wilie,
  Holy Lovenia, Ziwei Ji, Tiezheng Yu, Willy Chung, et~al. 2023.
\newblock A multitask, multilingual, multimodal evaluation of chatgpt on
  reasoning, hallucination, and interactivity.
\newblock \emph{arXiv preprint arXiv:2302.04023}.

\bibitem[{Brown et~al.(2020)Brown, Mann, Ryder, Subbiah, Kaplan, Dhariwal,
  Neelakantan, Shyam, Sastry, Askell et~al.}]{brown2020language}
Tom Brown, Benjamin Mann, Nick Ryder, Melanie Subbiah, Jared~D Kaplan, Prafulla
  Dhariwal, Arvind Neelakantan, Pranav Shyam, Girish Sastry, Amanda Askell,
  et~al. 2020.
\newblock Language models are few-shot learners.
\newblock In \emph{NeurIPs}, volume~33, pages 1877--1901.

\bibitem[{Bu et~al.(2023)Bu, Sheng, Cao, Qi, Wang, and Li}]{bu2023online}
Yuyan Bu, Qiang Sheng, Juan Cao, Peng Qi, Danding Wang, and Jintao Li. 2023.
\newblock Online misinformation video detection: a survey.
\newblock \emph{arXiv preprint arXiv:2302.03242}.

\bibitem[{Fan et~al.(2020)Fan, Piktus, Petroni, Wenzek, Saeidi, Vlachos,
  Bordes, and Riedel}]{fan2020generating}
Angela Fan, Aleksandra Piktus, Fabio Petroni, Guillaume Wenzek, Marzieh Saeidi,
  Andreas Vlachos, Antoine Bordes, and Sebastian Riedel. 2020.
\newblock Generating fact checking briefs.
\newblock In \emph{EMNLP}, pages 7147--7161.

\bibitem[{Gad-Elrab et~al.(2019)Gad-Elrab, Stepanova, Urbani, and
  Weikum}]{gad2019exfakt}
Mohamed~H Gad-Elrab, Daria Stepanova, Jacopo Urbani, and Gerhard Weikum. 2019.
\newblock Exfakt: A framework for explaining facts over knowledge graphs and
  text.
\newblock In \emph{WSDM}, pages 87--95.

\bibitem[{Hayes and Krippendorff(2007)}]{hayes2007answering}
Andrew~F Hayes and Klaus Krippendorff. 2007.
\newblock Answering the call for a standard reliability measure for coding
  data.
\newblock \emph{Communication methods and measures}, 1(1):77--89.

\bibitem[{Holtzman et~al.(2019)Holtzman, Buys, Du, Forbes, and
  Choi}]{holtzman2019curious}
Ari Holtzman, Jan Buys, Li~Du, Maxwell Forbes, and Yejin Choi. 2019.
\newblock The curious case of neural text degeneration.
\newblock In \emph{ICLR}.

\bibitem[{Ji et~al.(2023)Ji, Lee, Frieske, Yu, Su, Xu, Ishii, Bang, Madotto,
  and Fung}]{ji2023survey}
Ziwei Ji, Nayeon Lee, Rita Frieske, Tiezheng Yu, Dan Su, Yan Xu, Etsuko Ishii,
  Ye~Jin Bang, Andrea Madotto, and Pascale Fung. 2023.
\newblock Survey of hallucination in natural language generation.
\newblock \emph{ACM Computing Surveys}, 55(12):1--38.

\bibitem[{Jiang et~al.(2023)Jiang, Xu, Gao, Sun, Liu, Dwivedi-Yu, Yang, Callan,
  and Neubig}]{jiang2023active}
Zhengbao Jiang, Frank~F Xu, Luyu Gao, Zhiqing Sun, Qian Liu, Jane Dwivedi-Yu,
  Yiming Yang, Jamie Callan, and Graham Neubig. 2023.
\newblock Active retrieval augmented generation.
\newblock \emph{arXiv preprint arXiv:2305.06983}.

\bibitem[{Jin et~al.(2022)Jin, Wang, Yang, Sun, Wang, Liao, and
  Xie}]{jin2022towards}
Yiqiao Jin, Xiting Wang, Ruichao Yang, Yizhou Sun, Wei Wang, Hao Liao, and Xing
  Xie. 2022.
\newblock Towards fine-grained reasoning for fake news detection.
\newblock In \emph{AAAI}, volume~36, pages 5746--5754.

\bibitem[{Kenton and Toutanova(2019)}]{kenton2019bert}
Jacob Devlin Ming-Wei~Chang Kenton and Lee~Kristina Toutanova. 2019.
\newblock Bert: Pre-training of deep bidirectional transformers for language
  understanding.
\newblock In \emph{NAACL}, pages 4171--4186.

\bibitem[{Kotonya and Toni(2020)}]{kotonya2020explainable}
Neema Kotonya and Francesca Toni. 2020.
\newblock Explainable automated fact-checking for public health claims.
\newblock In \emph{EMNLP}, pages 7740--7754.

\bibitem[{Kruengkrai et~al.(2021)Kruengkrai, Yamagishi, and
  Wang}]{kruengkrai2021multi}
Canasai Kruengkrai, Junichi Yamagishi, and Xin Wang. 2021.
\newblock A multi-level attention model for evidence-based fact checking.
\newblock In \emph{ACL}, pages 2447--2460.

\bibitem[{Lazer et~al.(2018)Lazer, Baum, Benkler, Berinsky, Greenhill, Menczer,
  Metzger, Nyhan, Pennycook, Rothschild et~al.}]{lazer2018science}
David~MJ Lazer, Matthew~A Baum, Yochai Benkler, Adam~J Berinsky, Kelly~M
  Greenhill, Filippo Menczer, Miriam~J Metzger, Brendan Nyhan, Gordon
  Pennycook, David Rothschild, et~al. 2018.
\newblock The science of fake news.
\newblock \emph{Science}, 359(6380):1094--1096.

\bibitem[{Lee et~al.(2021)Lee, Bang, Madotto, and Fung}]{lee2021towards}
Nayeon Lee, Yejin Bang, Andrea Madotto, and Pascale Fung. 2021.
\newblock Towards few-shot fact-checking via perplexity.
\newblock In \emph{NAACL}, pages 1971--1981.

\bibitem[{Lee et~al.(2020)Lee, Li, Wang, Yih, Ma, and Khabsa}]{lee2020language}
Nayeon Lee, Belinda~Z Li, Sinong Wang, Wen-tau Yih, Hao Ma, and Madian Khabsa.
  2020.
\newblock Language models as fact checkers?
\newblock In \emph{FEVER Workshop}, pages 36--41.

\bibitem[{Lewis et~al.(2020)Lewis, Perez, Piktus, Petroni, Karpukhin, Goyal,
  K{\"u}ttler, Lewis, Yih, Rockt{\"a}schel et~al.}]{lewis2020retrieval}
Patrick Lewis, Ethan Perez, Aleksandra Piktus, Fabio Petroni, Vladimir
  Karpukhin, Naman Goyal, Heinrich K{\"u}ttler, Mike Lewis, Wen-tau Yih, Tim
  Rockt{\"a}schel, et~al. 2020.
\newblock Retrieval-augmented generation for knowledge-intensive nlp tasks.
\newblock In \emph{NeurIPS}, volume~33, pages 9459--9474.

\bibitem[{Lu and Li(2020)}]{lu2020gcan}
Yi-Ju Lu and Cheng-Te Li. 2020.
\newblock Gcan: Graph-aware co-attention networks for explainable fake news
  detection on social media.
\newblock In \emph{ACL}, pages 505--514.

\bibitem[{Ma et~al.(2019)Ma, Gao, Joty, and Wong}]{ma2019sentence}
Jing Ma, Wei Gao, Shafiq Joty, and Kam-Fai Wong. 2019.
\newblock Sentence-level evidence embedding for claim verification with
  hierarchical attention networks.
\newblock In \emph{ACL}.

\bibitem[{Ousidhoum et~al.(2022)Ousidhoum, Yuan, and
  Vlachos}]{ousidhoum2022varifocal}
Nedjma Ousidhoum, Zhangdie Yuan, and Andreas Vlachos. 2022.
\newblock Varifocal question generation for fact-checking.
\newblock \emph{arXiv preprint arXiv:2210.12400}.

\bibitem[{Popat et~al.(2018)Popat, Mukherjee, Yates, and
  Weikum}]{popat2018declare}
Kashyap Popat, Subhabrata Mukherjee, Andrew Yates, and Gerhard Weikum. 2018.
\newblock Declare: Debunking fake news and false claims using evidence-aware
  deep learning.
\newblock In \emph{EMNLP}, pages 22--32.

\bibitem[{Press et~al.(2022)Press, Zhang, Min, Schmidt, Smith, and
  Lewis}]{press2022measuring}
Ofir Press, Muru Zhang, Sewon Min, Ludwig Schmidt, Noah~A Smith, and Mike
  Lewis. 2022.
\newblock Measuring and narrowing the compositionality gap in language models.
\newblock \emph{arXiv preprint arXiv:2210.03350}.

\bibitem[{Radford et~al.(2019)Radford, Wu, Child, Luan, Amodei, Sutskever
  et~al.}]{radford2019language}
Alec Radford, Jeffrey Wu, Rewon Child, David Luan, Dario Amodei, Ilya
  Sutskever, et~al. 2019.
\newblock Language models are unsupervised multitask learners.
\newblock \emph{OpenAI blog}, 1(8):9.

\bibitem[{Rashkin et~al.(2017)Rashkin, Choi, Jang, Volkova, and
  Choi}]{rashkin2017truth}
Hannah Rashkin, Eunsol Choi, Jin~Yea Jang, Svitlana Volkova, and Yejin Choi.
  2017.
\newblock Truth of varying shades: Analyzing language in fake news and
  political fact-checking.
\newblock In \emph{EMNLP}, pages 2931--2937.

\bibitem[{Silva et~al.(2021)Silva, Luo, Karunasekera, and
  Leckie}]{silva2021embracing}
Amila Silva, Ling Luo, Shanika Karunasekera, and Christopher Leckie. 2021.
\newblock Embracing domain differences in fake news: Cross-domain fake news
  detection using multi-modal data.
\newblock In \emph{AAAI}, volume~35, pages 557--565.

\bibitem[{Soleimani et~al.(2020)Soleimani, Monz, and
  Worring}]{soleimani2020bert}
Amir Soleimani, Christof Monz, and Marcel Worring. 2020.
\newblock Bert for evidence retrieval and claim verification.
\newblock In \emph{ECIR}, pages 359--366.

\bibitem[{Thorne et~al.(2018)Thorne, Vlachos, Christodoulopoulos, and
  Mittal}]{thorne2018fever}
James Thorne, Andreas Vlachos, Christos Christodoulopoulos, and Arpit Mittal.
  2018.
\newblock Fever: a large-scale dataset for fact extraction and verification.
\newblock In \emph{NAACL}, pages 809--819.

\bibitem[{Wang et~al.(2023)Wang, Xu, Lan, Hu, Lan, Lee, and Lim}]{wang2023plan}
Lei Wang, Wanyu Xu, Yihuai Lan, Zhiqiang Hu, Yunshi Lan, Roy Ka-Wei Lee, and
  Ee-Peng Lim. 2023.
\newblock Plan-and-solve prompting: Improving zero-shot chain-of-thought
  reasoning by large language models.
\newblock \emph{arXiv preprint arXiv:2305.04091}.

\bibitem[{Wang(2017)}]{wang2017liar}
William~Yang Wang. 2017.
\newblock “liar, liar pants on fire”: A new benchmark dataset for fake news
  detection.
\newblock In \emph{ACL (short)}, pages 422--426.

\bibitem[{Wang et~al.(2018)Wang, Ma, Jin, Yuan, Xun, Jha, Su, and
  Gao}]{wang2018eann}
Yaqing Wang, Fenglong Ma, Zhiwei Jin, Ye~Yuan, Guangxu Xun, Kishlay Jha, Lu~Su,
  and Jing Gao. 2018.
\newblock Eann: Event adversarial neural networks for multi-modal fake news
  detection.
\newblock In \emph{Proceedings of the 24th acm sigkdd international conference
  on knowledge discovery \& data mining}, pages 849--857.

\bibitem[{Wei et~al.(2022)Wei, Wang, Schuurmans, Bosma, Chi, Le, and
  Zhou}]{wei2022chain}
Jason Wei, Xuezhi Wang, Dale Schuurmans, Maarten Bosma, Ed~Chi, Quoc Le, and
  Denny Zhou. 2022.
\newblock Chain of thought prompting elicits reasoning in large language
  models.
\newblock \emph{arXiv preprint arXiv:2201.11903}.

\bibitem[{Wu et~al.(2019)Wu, Morstatter, Carley, and
  Liu}]{wu2019misinformation}
Liang Wu, Fred Morstatter, Kathleen~M Carley, and Huan Liu. 2019.
\newblock Misinformation in social media: definition, manipulation, and
  detection.
\newblock In \emph{SIGKDD}, volume~21, pages 80--90.

\bibitem[{Yang et~al.(2019)Yang, Pentyala, Mohseni, Du, Yuan, Linder, Ragan,
  Ji, and Hu}]{yang2019xfake}
Fan Yang, Shiva~K Pentyala, Sina Mohseni, Mengnan Du, Hao Yuan, Rhema Linder,
  Eric~D Ragan, Shuiwang Ji, and Xia Hu. 2019.
\newblock Xfake: Explainable fake news detector with visualizations.
\newblock In \emph{WWW}, pages 3600--3604.

\bibitem[{Yang et~al.(2018)Yang, Qi, Zhang, Bengio, Cohen, Salakhutdinov, and
  Manning}]{yang2018hotpotqa}
Zhilin Yang, Peng Qi, Saizheng Zhang, Yoshua Bengio, William Cohen, Ruslan
  Salakhutdinov, and Christopher~D Manning. 2018.
\newblock Hotpotqa: A dataset for diverse, explainable multi-hop question
  answering.
\newblock In \emph{EMNLP}, pages 2369--2380.

\bibitem[{Yang et~al.(2022)Yang, Ma, Chen, Lin, Luo, and
  Chang}]{yang2022coarse}
Zhiwei Yang, Jing Ma, Hechang Chen, Hongzhan Lin, Ziyang Luo, and Yi~Chang.
  2022.
\newblock A coarse-to-fine cascaded evidence-distillation neural network for
  explainable fake news detection.
\newblock In \emph{COLING}, pages 2608--2621.

\bibitem[{Yao et~al.(2023)Yao, Zhao, Yu, Du, Shafran, Narasimhan, and
  Cao}]{yao2023react}
Shunyu Yao, Jeffrey Zhao, Dian Yu, Nan Du, Izhak Shafran, Karthik Narasimhan,
  and Yuan Cao. 2023.
\newblock {ReAct}: Synergizing reasoning and acting in language models.
\newblock In \emph{ICLR}.

\bibitem[{Zhang et~al.(2023)Zhang, Li, Cui, Cai, Liu, Fu, Huang, Zhao, Zhang,
  Chen et~al.}]{zhang2023siren}
Yue Zhang, Yafu Li, Leyang Cui, Deng Cai, Lemao Liu, Tingchen Fu, Xinting
  Huang, Enbo Zhao, Yu~Zhang, Yulong Chen, et~al. 2023.
\newblock Siren's song in the ai ocean: A survey on hallucination in large
  language models.
\newblock \emph{arXiv preprint arXiv:2309.01219}.

\bibitem[{Zhou et~al.(2022)Zhou, Sch{\"a}rli, Hou, Wei, Scales, Wang,
  Schuurmans, Bousquet, Le, and Chi}]{zhou2022least}
Denny Zhou, Nathanael Sch{\"a}rli, Le~Hou, Jason Wei, Nathan Scales, Xuezhi
  Wang, Dale Schuurmans, Olivier Bousquet, Quoc Le, and Ed~Chi. 2022.
\newblock Least-to-most prompting enables complex reasoning in large language
  models.
\newblock \emph{arXiv preprint arXiv:2205.10625}.

\end{thebibliography}
\bibliographystyle{acl_natbib}
\newpage

\appendix

\section{Case Study}
Table~\ref{tbl:case} shows the example from LIAR where CoT 1) generates sentences containing fact hallucination, with inaccurate percentages (i.e., 57 and 1 percent), and 2) omits necessary verification of the given percentage. While its final prediction is correct, it still encounters the issue of omitting necessary thoughts by not verifying the authenticity of the ``1 percentage to food and agriculture'' claim before making a judgment on the entire claim's correctness. In contrast, our HiSS method examines each component of the claim, along with leveraging search engine to gather comprehensive information. However, it is worth noting that even though HiSS can generate a more detailed reasoning process, it still faces challenges related to repeated generations.
\begin{table*}[t!]
\centering
\small
\begin{tabular}{p{15cm}}
\toprule
\textbf{Ground-truth}: \sethlcolor{pink}\hl{False}. \\\textbf{Claim}: Says 57 percent of federal spending goes to the military and just 1 percent goes to food and agriculture, including food stamps.\\
\midrule
\textbf{Standard Prompting}\\ \sethlcolor{pink}\hl{True.}\\
\midrule
\textbf{Vanilla CoT}\\ According to the Congressional Budget Office, the federal government spent \$4.1 trillion. \sethlcolor{c3}\hl{Of that, 57 percent went to the military and 1 percent went to food and agriculture, including food stamps.} Thus the claim is classified as \sethlcolor{pink}\hl{true.}\\
\midrule
\textbf{ReAct} \\
\textit{Thought 1}: I need to search for information about federal spending to verify the claim that 57 percent goes to the military \\ \quad and 1 percent goes to food and agriculture, including food stamps.\\
\textit{Action 1}: Search[federal spending breakdown]\\
\textit{Observation 1}: This means federal spending was equal to 25\%  \\ \quad of the total gross domestic product (GDP), or economic ..."\\
\textit{Thought 2}: The observation contradicts the claim that 57 percent goes to the military and \sethlcolor{c3}\hl{ 1 percent to food and\\ \quad }  \sethlcolor{c3}\hl{ agriculture, including food stamps.} The claim appears to be false.\\
\textit{Action 2}: Finish[\sethlcolor{pink}\hl{False}]\\
\midrule
\textbf{HiSS}\\
A fact checker will decompose the claim into 2 subclaims that are easier to verify:\\
\quad 1. 57 percent of federal spending goes to the military.\\
\quad 2. Just 1 percent of federal spending goes to food and agriculture, including food stamps.\\
To verify subclaim 1, a fact-checker will go through a step-by-step process to ask and answer a series of questions relevant to its factuality. Here are the specific steps he/she raise each question and look for an answer:\\
\quad Question: What percentage of federal spending goes to the military? \\
\quad Tell me if you are confident to answer the question or not. Answer with ``yes'' or ``no'': Yes.\\
\quad Answer: About one-sixth of federal spending goes to national defense. CBO estimates the budgetary effects of\\ \quad legislation related to national security and assesses the cost-effectiveness of current and proposed defense programs.\\\quad  CBO also analyzes federal programs and issues related to veterans.\\
To verify subclaim 2, a fact-checker will go through a step-by-step process to ask and answer a series of questions relevant to its factuality. Here are the specific steps he/she raise each question and look for an answer:\\
\quad Question: What percentage of federal spending goes to food and agriculture, including food stamps? \sethlcolor{c2}\hl{Question: What}\\ \quad \sethlcolor{c2}\hl{percentage of federal spending goes to food and agriculture, including food stamps? }\\
 \quad Tell me if you are confident to answer the question or not. Answer with ``yes'' or ``no'': No.\\
\quad Answer: Federal spending on USDA's food and nutrition assistance programs totaled \$182.5 billion, 49 percent more \\ \quad than the ... USDA's food and nutrition assistance programs accounted for about 5 percent of total federal outlays.\\
\sethlcolor{c2}\hl{Based on the answers to these questions, it is clear that among pants-fire, false, barely-true, half-true, mostly-true, and true, the claim can be classified as false.} Based on the answers to these questions, it is clear that among pants-fire, false, barely-true, half-true, mostly-true, and true, the claim can be classified as \sethlcolor{pink}\hl{false}.\\
\bottomrule

\end{tabular}
\caption{Case study. Labels and predictions from different methods are in pink, generated sentences containing fact hallucination are in purple, and repeated sentences are in yellow. }
\label{tbl:case}

\end{table*}

\section{Demonstration Examples}
\label{app:prompt}
We show the demonstration examples used in the LIAR dataset.
Table~\ref{tbl:sp} and~\ref{tbl:cot} present the prompts we used for standard prompting and vanilla CoT, respectively. Table~\ref{tbl:hiss} and~\ref{tbl:hiss_2} present the prompts used for HiSS.

\begin{table*}[t!]
\centering
\small
\begin{tabular}{p{15cm}}
\toprule
\textbf{Q:}
Among pants-fire, false, barely-true, half-true, mostly-true, and true, the claim "Emerson Moser, who was Crayola’s top crayon molder for almost 40 years, was colorblind." is classified as \\
\textbf{A:}
mostly-true.\\\\
\textbf{Q:}
Among pants-fire, false, barely-true, half-true, mostly-true, and true, the claim "Bernie Sanders said 85 million Americans have no health insurance." is classified as \\
\textbf{A:}
half-true.\\\\
\textbf{Q:}
Among pants-fire, false, barely-true, half-true, mostly-true, and true, the claim "JAG charges Nancy Pelosi with treason and seditious conspiracy." is classified as \\
\textbf{A:}
pants-fire.\\\\
\textbf{Q:}
Among pants-fire, false, barely-true, half-true, mostly-true, and true, the claim "Cheri Beasley “backs tax hikes — even on families making under \$75,000." is classified as \\\textbf{A:}barely-true.\\
\bottomrule

\end{tabular}
\caption{Demonstration examples of standard prompting.  }
\label{tbl:sp}

\end{table*}
\begin{table*}[t!]
\centering
\small
\begin{tabular}{p{15cm}}
\toprule
\textbf{Q:} Choose a label from pants-fire, false, barely-true, half-true, mostly-true, and true for the following claim.\\
Claim: "Emerson Moser, who was Crayola’s top crayon molder for almost 40 years, was colorblind." \\
\textbf{A:} \sethlcolor{c1}\hl{Moser did have a type of colorblindness. However he said he only had blue-green colorblindness, So only some colors were confusing to him.} Thus the claim is classified as mostly-true.\\\\
\textbf{Q:} Choose a label from pants-fire, false, barely-true, half-true, mostly-true, and true for the following claim.\\
Claim:  "Bernie Sanders said 85 million Americans have no health insurance."\\
\textbf{A:} \sethlcolor{c1}\hl{Sanders accurately cited the Commonwealth Fund report in an October op-ed. His statement on CNN was partially accurate but left out important details and context. Sanders omitted that his figure included people who either have no health insurance or are underinsured.} Thus the claim is classified as half-true.\\\\
\textbf{Q:} Choose a label from pants-fire, false, barely-true, half-true, mostly-true, and true for the following claim.\\
Claim:  "JAG charges Nancy Pelosi with treason and seditious conspiracy."\\
\textbf{A:} \sethlcolor{c1}\hl{A spokesperson for the U.S. Navy Judge Advocate General's Corps told PolitiFact it isn't true that Pelosi has been charged. The original claim appeared on a website that posts fantastical, false stories and has a disclaimer that the\\ site contains ``humor, parody, and satire.''} Thus the claim is classified as pants-fire.\\\\
\textbf{Q:} Choose a label from pants-fire, false, barely-true, half-true, mostly-true, and true for the following claim.\\
Claim:  "Cheri Beasley “backs tax hikes — even on families making under \$75,000."\\
\textbf{A:} \sethlcolor{c1}\hl{Beasley does support the Inflation Reduction Act but the claim is wrong to suggest it includes a change in tax rates for that income bracket. Expert analyses show that any effect on those earners would be on the scale of \$350 and, by one estimate, delayed for several years.} Thus the claim is classified as barely-true.\\

\bottomrule

\end{tabular}
\caption{Demonstration examples of vanilla CoT prompting. The CoT process is highlighted in green. }
\label{tbl:cot}

\end{table*}
\begin{table*}[t!]
\renewcommand{\thetable}{8 (a)}
\centering
\small
\begin{tabular}{p{15cm}}
\toprule
\textbf{Q:} Claim: ``Emerson Moser, who was Crayola’s top crayon molder for almost 40 years, was colorblind.''\\
\textbf{A:} A fact checker will decompose the claim into 4 subclaims that are easier to verify:\\
 \quad \sethlcolor{c1}\hl{1. Emerson Moser was a crayon molder at Crayola.}\\
 \quad \sethlcolor{c1}\hl{2. Moser worked at Crayola for almost 40 years.}\\
 \quad \sethlcolor{c1}\hl{3. Moser was Crayola's top crayon molder.}\\
 \quad \sethlcolor{c1}\hl{4. Moser was colorblind.}\\
To verify subclaim 1, a fact-checker will go through a step-by-step process to ask and answer a series of questions relevant to its factuality. Here are the specific steps he/she raise each question and look for an answer:\\
 \quad  \sethlcolor{c1}\hl{Question: Is there any official record or documentation indicating that Emerson Moser worked as a crayon molder}\\  \quad  \sethlcolor{c1}\hl{at Crayola?}\\
 \quad Tell me if you are confident to answer the question or not. Answer with ``yes'' or ``no'': Yes.\\
 \quad  \sethlcolor{c1}\hl{Answer: Yes.}\\
To verify subclaim 2, a fact-checker will go through a step-by-step process to ask and answer a series of questions relevant to its factuality. Here are the specific steps he/she raise each question and look for an answer:\\
 \quad  \sethlcolor{c1}\hl{Question: Are there any official records or documentation confirming Emerson Moser's length of employment at}\\   \sethlcolor{c1}\hl{\quad Crayola?}\\
 \quad Tell me if you are confident to answer the question or not. Answer with ``yes'' or ``no'': No.\\
 \quad  \sethlcolor{c1}\hl{Answer: Emerson Moser, who is retiring next week after 35 years, isn't colorblind in the sense that he can't see}\\   \sethlcolor{c1}\hl{\quad color at all. It's just that some ...}\\
To verify subclaim 3, a fact-checker will go through a step-by-step process to ask and answer a series of questions relevant to its factuality. Here are the specific steps he/she raise each question and look for an answer:\\
 \quad  \sethlcolor{c1}\hl{Question: Are there credible sources or publications that mention Emerson Moser as Crayola's top crayon molder?}\\
 \quad Tell me if you are confident to answer the question or not. Answer with ``yes'' or ``no'': Yes.\\
 \quad  \sethlcolor{c1}\hl{Answer: Yes.}\\
To verify subclaim 4, a fact-checker will go through a step-by-step process to ask and answer a series of questions relevant to its factuality. Here are the specific steps he/she raise each question and look for an answer:\\
 \quad  \sethlcolor{c1}\hl{Question: Are there any credible sources or records indicating that Emerson Moser was colorblind?}\\
 \quad Tell me if you are confident to answer the question or not. Answer with ``yes'' or ``no'': Yes.\\
 \quad  \sethlcolor{c1}\hl{Answer: Yes.}\\
 \quad  \sethlcolor{c1}\hl{Question: Was Emerson Moser's colorblindness only confusing for certain colors?} \\
 \quad Tell me if you are confident to answer the question or not. Answer with ``yes'' or ``no'': No.\\
 \quad  \sethlcolor{c1}\hl{Answer: Moser has had tritanomaly, a type of colorblindness that makes it difficult to distinguish between blue and}\\  \quad  \sethlcolor{c1}\hl{green and between yellow and red.}\\
Based on the answers to these questions, it is clear that among pants-fire, false, barely-true, half-true, mostly-true, and true, the claim can be classified as mostly-true.\\\\

\textbf{Q:} Claim: ``Bernie Sanders said 85 million Americans have no health insurance.''\\
\textbf{A:} A fact checker will not split the claim since the original claim is easier to verify.\\
To verify the claim, a fact-checker will go through a step-by-step process to ask and answer a series of questions relevant to its factuality. Here are the specific steps he/she raise each question and look for an answer:\\
\quad  \sethlcolor{c1}\hl{Question: How many Americans did Bernie Sanders claim had no health insurance?}\\
\quad Tell me if you are confident to answer the question or not. Answer with ``yes'' or ``no'': No.\\
\quad  \sethlcolor{c1}\hl{Answer: ``We have 85 million Americans who have no health insurance,'' Sanders said Dec. 11 on CNN's State of}\\  \quad  \sethlcolor{c1}\hl{the Union.}\\
\quad  \sethlcolor{c1}\hl{Question: How did Bernie Sanders define ``no health insurance''?}\\
\quad Tell me if you are confident to answer the question or not. Answer with ``yes'' or ``no'': No.\\
\quad  \sethlcolor{c1}\hl{Answer: Sanders spokesperson Mike Casca said the senator was referring to the number of uninsured and}\\  \quad  \sethlcolor{c1}\hl{under-insured Americans and cited a report about those numbers for adults.}\\
\quad  \sethlcolor{c1}\hl{Question: How many Americans were uninsured or under-insured according}\\  \quad  \sethlcolor{c1}\hl{to the Commonwealth Fund survey?}\\
\quad Tell me if you are confident to answer the question or not. Answer with ``yes'' or ``no'': No.\\
\quad  \sethlcolor{c1}\hl{Answer: The Commonwealth Fund survey found that 43\% of working-age adults 19 to 64, or about 85 million}\\ \quad  \sethlcolor{c1}\hl{Americans, were uninsured or inadequately insured.} \\
\quad  \sethlcolor{c1}\hl{Question: Is the statement ``we have 85 million Americans who have no health insurance'' partially accurate according}\\ \quad  \sethlcolor{c1}\hl{to the information in the passage?}\\
\quad Tell me if you are confident to answer the question or not. Answer with ``yes'' or ``no'': No.\\
\quad  \sethlcolor{c1}\hl{Answer: Bernie Sanders omitted that his figure included people who either have no health insurance or are}\\  \quad  \sethlcolor{c1}\hl{under-insured.}\\
Based on the answers to these questions, it is clear that among pants-fire, false, barely-true, half-true, mostly-true, and true, the claim is classified as half-true.\\
\bottomrule

\end{tabular}
\caption{Demonstration examples of HiSS prompting. The CoT process is highlighted in green.  }
\label{tbl:hiss}

\end{table*}
\begin{table*}[t!]
\renewcommand{\thetable}{8 (b)}
\centering
\small
\begin{tabular}{p{15cm}}
\toprule
\textbf{Q:} Claim: ``JAG charges Nancy Pelosi with treason and seditious conspiracy.''\\
\textbf{A:} A fact checker will decompose the claim into 2 subclaims that are easier to verify:\\
\quad  \sethlcolor{c1}\hl{1. JAG has made a claim or accusation against Nancy Pelosi.}\\
\quad  \sethlcolor{c1}\hl{2. The specific charges or allegations made against Nancy Pelosi are treason and seditious conspiracy.}\\
To verify subclaim 1, a fact-checker will go through a step-by-step process to ask and answer a series of questions relevant to its factuality. Here are the specific steps he/she raise each question and look for an answer:\\
\quad  \sethlcolor{c1}\hl{Question: Is it true that JAG has made a claim or accusation against Nancy Pelosi?}\\
\quad Tell me if you are confident to answer the question or not. Answer with ``yes'' or ``no'': No.\\
\quad  \sethlcolor{c1}\hl{Answer: There is no evidence to support this claim and a spokesperson for the U.S. Navy Judge Advocate General's} \\\quad  \sethlcolor{c1}\hl{Corps has stated that it is not true.}\\ 
To verify subclaim 2, a fact-checker will go through a step-by-step process to ask and answer a series of questions relevant to its factuality. Here are the specific steps he/she raise each question and look for an answer:\\
\quad  \sethlcolor{c1}\hl{Question: Is it true that the specific charges or allegations made against Nancy Pelosi are treason and seditious} \\\quad  \sethlcolor{c1}\hl{conspiracy?}\\
\quad Tell me if you are confident to answer the question or not. Answer with ``yes'' or ``no'': No.\\
\quad  \sethlcolor{c1}\hl{Answer: There is no evidence to support this claim.}\\
\quad  \sethlcolor{c1}\hl{Question: Where is the source of the claim?}\\
\quad Tell me if you are confident to answer the question or not. Answer with ``yes'' or ``no'': No.\\
\quad  \sethlcolor{c1}\hl{Answer: Real Raw News, a disclaimer stating that it contains ``humor, parody and satire'' and has a history of} \\\quad  \sethlcolor{c1}\hl{publishing fictitious stories.} \\
Based on the answers to these questions, it is clear that among pants-fire, false, barely-true, half-true, mostly-true, and true, the claim is classified as pants-fire.\\\\

\textbf{Q:} Claim: ``Cheri Beasley “backs tax hikes — even on families making under \$75,000.''\\
\textbf{A:} A fact checker will decompose the claim into 2 subclaims that are easier to verify:\\
\quad  \sethlcolor{c1}\hl{1. Cheri Beasley supports tax increases.}\\
\quad  \sethlcolor{c1}\hl{2. Cheri Beasley supports tax increases for families with an income under \$75,000.}\\
To verify subclaim 1, a fact-checker will go through a step-by-step process to ask and answer a series of questions relevant to its factuality. Here are the specific steps he/she raise each question and look for an answer:\\
\quad  \sethlcolor{c1}\hl{Question: Does Cheri Beasley supports tax increases?}\\
\quad Tell me if you are confident to answer the question or not. Answer with ``yes'' or ``no'': Yes.\\
\quad  \sethlcolor{c1}\hl{Answer: Beasley supports student loan bailouts for the wealthy.}\\
To verify subclaim 2, a fact-checker will go through a step-by-step process to ask and answer a series of questions relevant to its factuality. Here are the specific steps he/she raise each question and look for an answer:\\
\quad  \sethlcolor{c1}\hl{Question: Does the ad accurately link Beasley's position on student loan debt forgiveness with her stance on tax hikes} \\\quad  \sethlcolor{c1}\hl{for families making under \$75,000 per year?}\\
\quad Tell me if you are confident to answer the question or not. Answer with ``yes'' or ``no'': No.\\
\quad  \sethlcolor{c1}\hl{Answer: The ad makes a misleading connection between the two issues and does not accurately represent Beasley's} \\\quad  \sethlcolor{c1}\hl{position on tax hikes for families making under \$75,000 per year.}\\
\quad  \sethlcolor{c1}\hl{Answer: No.}\\
Based on the answers to these questions, it is clear that among pants-fire, false, barely-true, half-true, mostly-true, and true, the claim is classified as barely-true.\\

\bottomrule

\end{tabular}
\caption{Demonstration examples of HiSS prompting. The CoT process is highlighted in green. }
\label{tbl:hiss_2}

\end{table*}

\end{document}